# Toward Multiphysics-Informed Machine Learning for Sustainable Data Center Operations

*Intelligence Evolution with Deployable Solutions for Computing Infrastructure*

Ruihang Wang, Qingang Zhang, Jimin Jia, Yonggang Wen, Stuart Kennedy

The revolution in artificial intelligence (AI) has brought sustainable challenges in data center management due to the high carbon emissions and short cooling response time associated with high-power density racks. While machine learning (ML) offers promise for intelligent management, its adoption is hindered by safety and reliability concerns. To address this, we propose a multiphysics-informed machine learning (MPIML) framework that integrates physical priors into data-driven models for enhanced accuracy and safety. We introduce an integrated system architecture comprising three core engines: DCLib for versatile facility modeling, DCTwin for high-fidelity multiphysics simulation, and DCBrain for decision-making optimization**.** This system enables critical predictive and prescriptive applications, such as carbon-aware IT provisioning, safety-aware intelligent cooling control and battery health forecasting. An illustrative example on an industry-grade data center cooling control demonstrates that our MPIML approach reduces annual carbon emissions up to 200 kilotons compared with conventional methods while ensuring operational constraints are met. We conclude by outlining key challenges and future directions for developing autonomous and sustainable data centers.

## 1. Introduction

The data center (DC) industry is experiencing rapid growth driven by the increasing demand for cloud computing, data storage, and artificial intelligence (AI) services. As DCs grow in scale and complexity, so does their carbon emissions. According to the International Energy Agency (IEA), global DC carbon emissions are projected to reach 320 megatons from 2025 to 2030, with a compound annual growth rate of 9.9%. This growth is primarily due to the rise of DCs dedicated for AI training and inference services. The statistics show that more recent large language models have significantly higher emissions for training, such as 588 tons for GPT-3, 5,184 tons for GPT-4, and 8,930 tons for Llama 3.1 (405B).

The growth of DC carbon emissions brings operational challenges in improving sustainability while ensuring system reliability. On the one hand, the amount of operational carbon emitted by a DC is determined by many factors, including the DC's power usage effectiveness (PUE) and the grid carbon intensity. Specifically, PUE is the ratio of the total amount of energy used by a DC, including cooling, to the energy delivered to IT equipment. Lower PUE indicates the DC is operated with higher energy efficiency. The grid carbon intensity is the amount of greenhouse gas emitted per unit of electricity supplied and varies by location and over time with the grid's generation mix. Lower intensity translates to lower operational emissions for the same energy use. Therefore, a key way to improve a DC's sustainability is to reduce its PUE and carbon intensity of energy supply, for example by cooling optimization or shifting IT workloads to lower-carbon times or regions. On the other hand, the reliability of a DC is measured by its uptime, which determines the business continuity. The reliability is related to cyber and physical factors. In this article, we focus on physical factors (e.g., temperature and humidity) that impact a DC's operational reliability.

To address the challenges, machine learning (ML)-based approaches have shown considerable promise. For example, recent study by Wei et al. has explored the use of deep reinforcement learning (DRL) to develop energy-efficient control policies for DC cooling systems by interacting with simulated DC environments. Compared to traditional feedback control systems, such as the commonly used proportional-integral-derivative (PID) controllers, DRL-based approaches offer advanced capabilities in predictive control and multi-objective optimization. They have demonstrated 26.9% energy and associated carbon emission reductions for cooling system over conventional feedback controllers. Despite these promising results, only 19% of individuals are confident in deploying ML-based solutions to a real-world DC. The low confidence is primarily due to two reasons. First, purely data-driven models often require large, diverse datasets, including the rare and abnormal cases that are costly to collect from a DC in stable operations. Without sufficient data coverage, ML models may produce physically implausible predictions when evaluated on data points not covered during training. Second, as a mission-critical infrastructure, the risk-aversion mindset curtails the implementation of ML-based policies in DCs. The ML-based policies may introduce risks without adequate safety validation.

To bridge the gaps, this article proposes to integrate the physics priors to advance traditional ML-based techniques. Physics-informed machine learning (PIML) is an emerging topic in applied machine learning. In addition to the training data, the physics priors aim to regulate the data-driven models to yield more accurate and physically plausible results. Given the potential benefits of incorporating physics into ML models, investigating PIML for DC modeling and optimization is a valuable pursuit. However, DCs are highly integrated, complex, and multidisciplinary cyber-physical systems (CPS), single-domain physics is insufficient to capture the holistic system dynamics. Moreover, from the standpoint of system optimization, optimizing a single domain does not guarantee the best outcome for the entire system. To this end, we propose a multiphysics-informed machine learning (MPIML) framework to advance DC operations. The MPIML-empowered solution has the potential to mitigate the data requirements for holistic system modeling and significantly accelerate simulation speed compared with traditional numerical techniques, making it suitable for various DC optimizations.

To present the MPIML-empowered solution, we first provide a reference framework to illustrate the data-physics availability in different stages of a DC, and further explore the role of physics in enhancing ML models in different subsystems. Then, we propose an integrated system architecture that consists of the DCLib, DCTwin, and DCBrain engines, to enable MPIML technology development and deployment for comprehensive DC modeling and sustainability optimization. An illustrative example on an industry-grade DC cooling optimization is presented to demonstrate the effectivenesss of the MPIML-empowered approach. With the proposed solution, we highlight the challenges and opportunities in physics-informed approaches to advance DC sustainability and shed light on future directions.

## 2. A Blueprint for the Physics-Aware DC

This section provides an overview of the interconnected DC system from a multiphysics perspective and illustrates the availability of data and physics in different modelling scenarios.

### 2.1 Deconstructing the DC: A Multiphysics System

DC is a complex CPS that involves multidisciplinary knowledge. Figure 1 illustrates the architecture of a typical DC that consists of three major interconnected subsystems, i.e., the computing system, the cooling system, and the power supply system. The interconnection of these systems involves the interplay of various physical processes, such as fluid dynamics, heat transfer, mechanical principles, electromagnetism, etc. To characterize the coupled dynamics of a DC, a comprehensive model of a DC must account for several key physical domains:

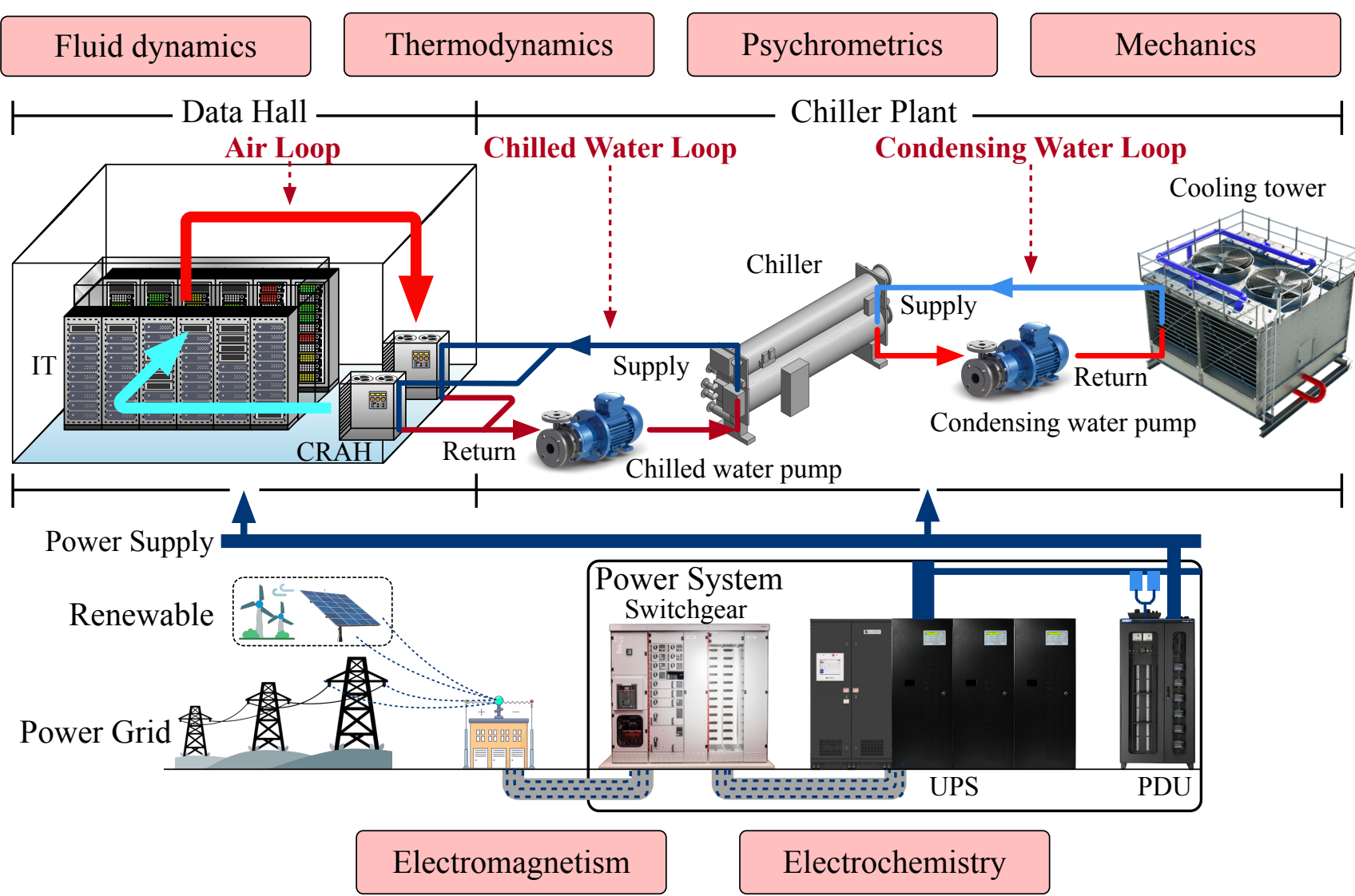

*Figure 1 A typical chilled water-cooled DC. It shows three interconnected systems: computing, cooling, and electrical supply. The cooling system is detailed with a data hall air loop, a chiller plant with a chilled water loop, and a condensing water loop. The power system includes the power grid, renewables, UPS, and PDU. Different physical processes are mapped to these components.*

**Thermodynamics and Heat Transfer**: This heat transfer process involves mechanical devices such as computer room air handling (CRAH) units or pumps to circulate fluid, which also consumes electricity. For a typical chilled-water cooled system, the heat is removed from the data hall via the indoor air cycle, the chilled water cycle, and the condensed water cycle, respectively.

**Fluid Dynamics**: This domain describes the motion of fluids within the facility. It includes the complex, often turbulent, airflow patterns in the data hall, shaped by factors like hot/cold aisle containment and underfloor plenums, as well as the flow of chilled water through the chiller plant's piping network and the circulation of refrigerant within CRAH units.

**Psychrometrics**: This specialized field of thermodynamics deals with the properties of gas-vapor mixtures, specifically the air and water vapor present in the data hall. Precise control of humidity is critical for preventing hardware damage from either corrosion with high humidity or electrostatic discharge with low humidity, making psychrometric modeling essential.

**Power and Energy**: This domain encompasses the electrical power consumption of every component in the facility, from servers and network switches to the pumps, fans, and compressors that drive the cooling system. The fundamental principle of energy conservation—that all electrical power consumed by the IT equipment is ultimately converted into heat that must be removed—provides a powerful constraining law for the entire system.

The physical priors we integrate are fundamental principles governing the DC's systems. For instance, in an air-cooled data hall, we need to consider thermal-fluid coupling by integrating the Navier-Stokes and energy balance equations to model the in-hall thermodynamics. For heat exchangers between air and chilled water loops, the energy balance principle should be incorporated to ensure that the heat generated by IT equipment, plus other gains, equals the heat removed by the cooling system. For mechanical system power usage, we use the affinity laws for pumps and fans, which provide a well-defined relationship between their speed, flow rate, and power consumption. For example, power is proportional to the cube of the speed. These established equations provide a strong physical foundation that guides the machine learning models. Moreover, the tight coupling between these domains means that an action taken in one can have significant and sometimes counterintuitive effects in another. For this reason, a multiphysics modeling approach is necessary, as optimizing a single domain, such as the data hall's airflow, does not guarantee an optimal outcome for the entire system.

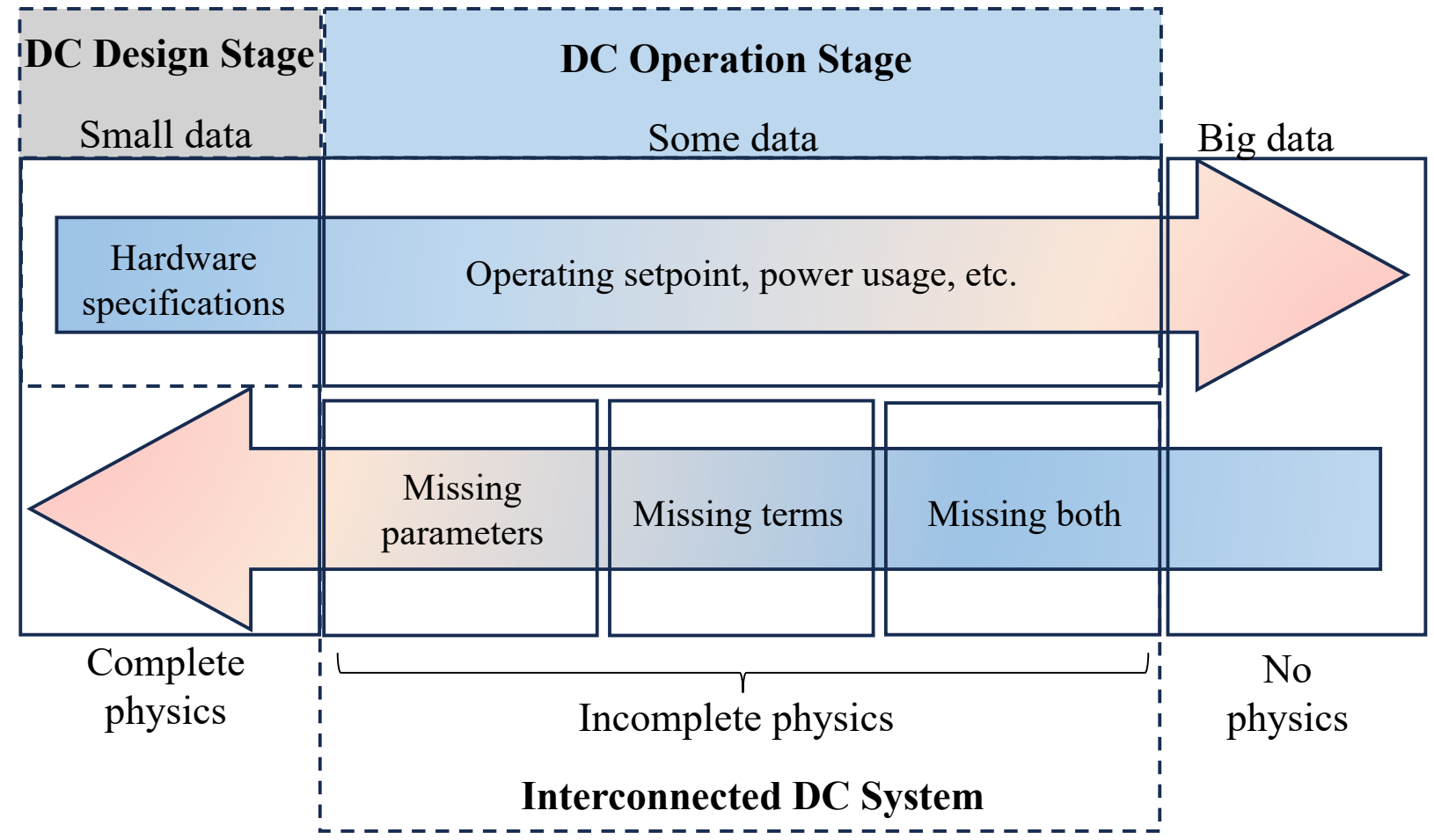

*Figure 2 The completeness of data and physics in different DC stages. The middle areas represent typical scenarios in a DC where physics are incomplete with either missing parameter values or unmodeled terms.*

## 2.2 Navigating the Data-Physics Spectrum

The success of modeling the complex DC system highly depends on the amount of available data and prior knowledge. The development of an MPIML model for a DC must be guided by a clear understanding of the available resources, which can be conceptualized along two axes, i.e., data availability and physics completeness as shown in Figure 2. Data availability ranges from a data-poor environment, such as the early design phase where only simulation data is available, to a data-rich environment in an operating facility with a dense network of real-time sensors. Physics completeness ranges from systems where the governing equations are fully known and parameterized, to systems with complex, unknown dynamics. The most common scenario in an operational DC lies in the "middle ground" of this spectrum, where some operational data are available and some physical principles are known. However, due to operational requirements, the data are confined to limited ranges, and the models remain incomplete, either lacking specific parameters or missing certain terms. We categorize this regime into two scenarios.

**Model with Unknown Parameters**: In this scenario, while the physics equation of a specific process is known, the equation parameters are missing. This area involves identifying these unknown parameters from measured data. For example, the affinity laws governing the power consumption of a fan are well-established physical principles (e.g., power is proportional to the cube of the fan speed). However, the specific coefficients of this relationship are unique to each fan model and installation and must be identified from operational data.

**Model with Missing Terms**: This scenario indicates that while the major physical process is modeled, the remaining factors are not precisely considered. For example, a thermodynamic model of a data hall might accurately capture the dominant heat source, the IT equipment, but neglect secondary heat sources like solar gain through windows, heat infiltration through walls, or the metabolic heat load from human operators. An ML model can be trained to learn a residual function that accounts for these unmodeled physical effects.

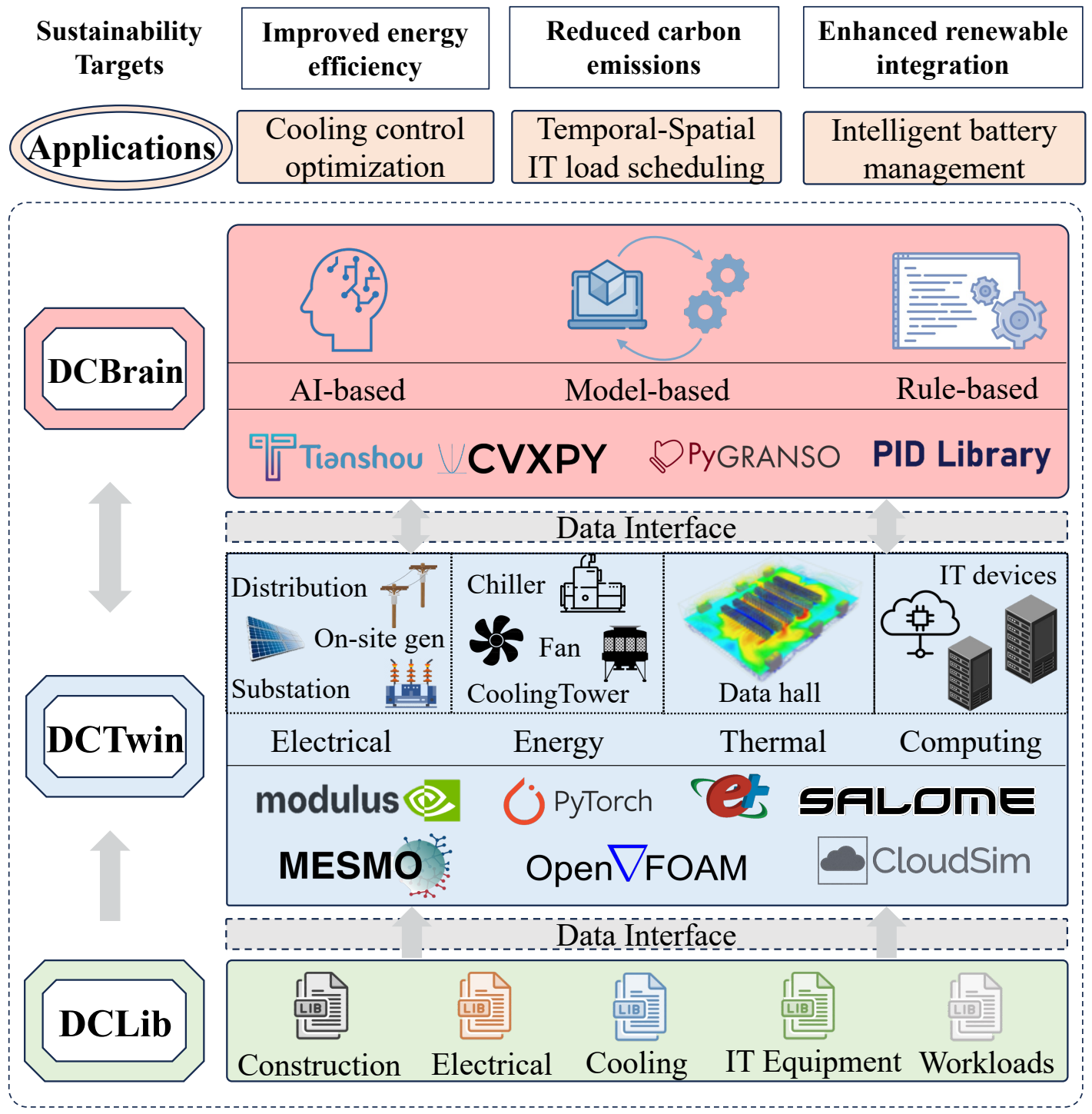

*Figure 3 Architecture of the core engines to empower PIML for DC modeling and optimization. The system consists of three engines for DC modeling and optimization, i.e., the DCLib, DCTwin, and DCBrain. The applications illustrate the potential of the proposed framework to tangible sustainability outcomes, such as improved energy efficiency/PUE, reduced carbon emissions, and enhanced integration of renewable energy.*

## 3. PIML System & Intelligence Evolution for Sustainable DC

This section presents the MPIML-empowered system architecture design and intelligence evolution for DC management.

### 3.1 An Integrated MPIML Architecture for DCs

Figure 3 shows the architecture design of the MPIML-empowered system for DC. It consists of three interconnected engines, i.e., the DCLib, DCTwin, and DCBrain, to support cross-domain modeling and optimization.

**DCLib**: We first develop a versatile Python-based library to facilitate the creation of facility models in DCs. The DCLib provides a holistic framework that encapsulates various aspects of DC objects, ranging from the cyber level to the physical level. Specifically, the library includes modules for construction, offering data classes for key building materials such as walls and floors. It also includes a cooling module, which provides classes for cooling systems like CRAH units, chillers, and cooling towers. Furthermore, the library covers the electrical infrastructure of DCs, including transformers and generators. The IT equipment module encapsulates classes for servers and racks, while the workload module handles various IT workloads including jobs and tasks. With its comprehensive modules, DCLib serves as a one-stop solution for building, managing, and simulating DC models.

**DCTwin**: DCTwin is an all-encompassing platform for high-fidelity, interconnected DC digital twin engine, allowing for sophisticated multiphysics simulations. It accepts the modeling configuration file from DCLib as the input. The key components of DCTwin include a differentiable physics-informed model base built on PyTorch and Nvidia PhysicsNeMo for various domain simulations, including thermodynamics, fluid dynamics, psychrometrics, and facility energy consumption. DCTwin also integrates a variety of open-source modeling techniques from diverse disciplines to provide training data and evaluate the performance of the PIML models. Key components include OpenFOAM for CFD modeling, EnergyPlus for facility power analysis, Mesmo for electrical distribution, and CloudSim for IT workload scheduling.

**DCBrain**: At the top of the system is the DCBrain engine tailored to various decision-making optimizations. This is achieved through the integration of various policies, including model-free, model-based, and rule-based approaches. These policies are developed based on state-of-the-art libraries, such as Tianshou, CVXPY, etc. Users can define optimization objectives and constraints related to DC operations and use the engine to derive the optimal policy, such as an energy-efficient DRL policy with safety considerations.

## 3.2 Intelligence Evolution for DC Management

Figure 4 illustrates the MPIML intelligence evolution that consists of three major development stages to advance DC operations. Specifically, the objectives of each stage are described as follows:

**Predictive Level Intelligence**: The foundational level of intelligence focuses on developing high-fidelity, real-time predictive models of the DC's state and the health of its components, including thermodynamics, facility energy consumption, equipment health status, etc. These predictive models can forecast changes across various boundary configurations and external inputs for what-if analyses. The prediction results can then be used to assess and validate potential system upgrades or degradation. To ensure high-fidelity prediction, this stage needs to calibrate unknown model parameters based on operational data. The calibrated models can subsequently be used to produce synthetic data to tackle the data scarcity issue as illustrated in §II. In some domains, the predictive models may demand significant computation resources for a single forward process, e.g., executing a CFD model for a fine-grained data hall temperature field prediction. The high overhead is inefficient in offline usage and prohibits the usage of online decision-making optimizations. To address this issue, effective model reduction is necessary to reduce the computation overhead of the original models. The reduced order models are expected to perform real-time prediction while maintaining satisfactory accuracy.

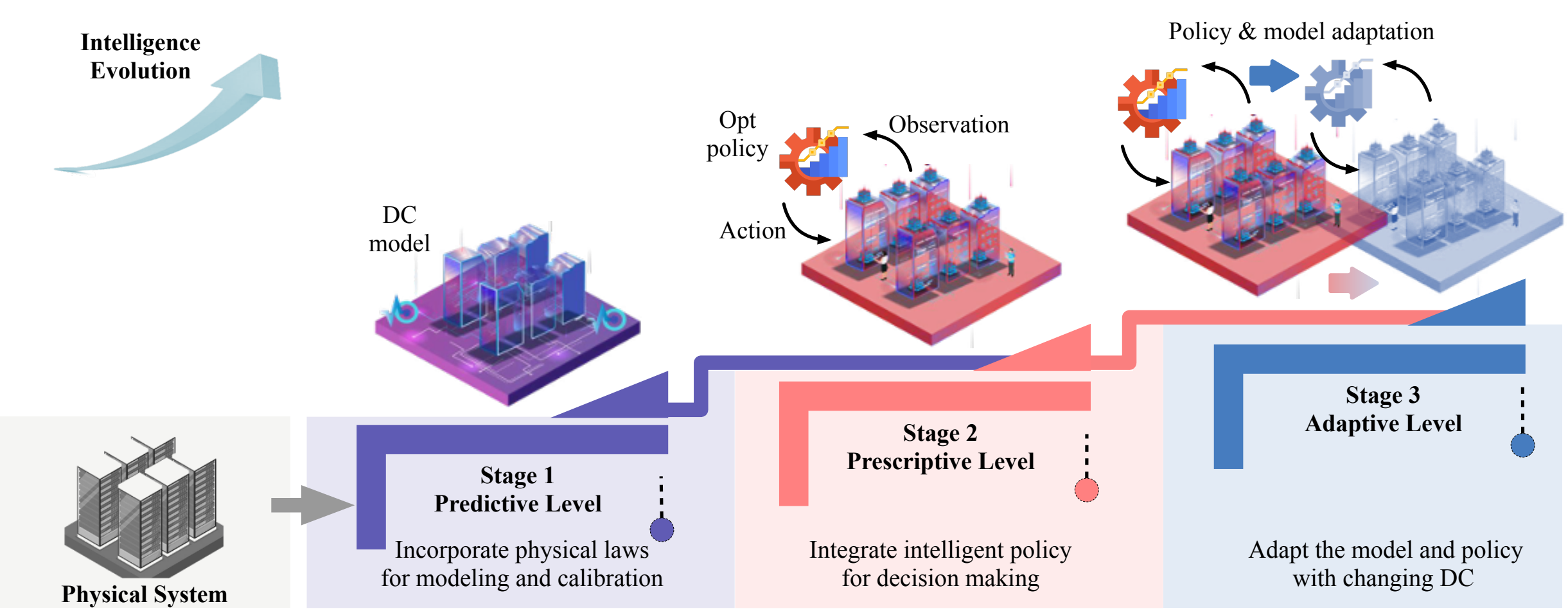

*Figure 4 MPIML intelligence evolution in three levels. The predictive level intelligence aims to perform accurate and timely predictions. The prescriptive level intelligence aims to facilitate decision-making with the previous developed models. Finally, the policies and models are deployed to physical DC with continuous adaptation.*

**Prescriptive Level Intelligence**: The second level of intelligence marks the transition from passive prediction to active, online decision-making and optimization. Besides offline usage for what-if analyses, the predictive models developed in the previous level can be used to facilitate online decision-making optimizations. This level aims to develop models with prescriptive capabilities to advance DC operations, such as energy-efficient cooling control, renewable-aware IT resources planning, battery system maintenance, etc. The predictive models can facilitate policy development via sampled-based or gradient-based methods. For sampled-based methods, DRL is often adopted to learn a parameterized policy guided by a reward function that incorporates multiple objectives. For example, compared with traditional rule-based control that focuses only on a single objective, such as maintaining the temperature at a predetermined setpoint, the DRL-based policy can incorporate both power usage and temperature into the reward function design. With the developed predictive models, the DRL-based policy can be trained via extensive model interaction. For gradient-based methods, the objective and constraint functions should be differentiable in terms of decision variables. To meet this requirement, the predictive models should be able to provide gradient information via automatic differentiation.

**Adaptive Level Intelligence**: At the top of the roadmap is the adaptive level intelligence, which functions throughout the entire DC lifecycle. After online deployment, this stage aims to provide a fully automated solution to adapt the previously developed models and policies with system upgrades, such as facility upgrades. Compared with the previous stages, the models and policies are expected to maintain satisfactory performance as system configurations change. To satisfy specific system constraints, this stage requires approaches that can facilitate fast and safe adaptations. Therefore, the previously developed predictive models are expected to extrapolate the changing system dynamics and configurations over time. In this respect, the MPIML-empowered solutions show promise at this level, as the fundamental physical principles of DC remain consistent throughout the DC lifecycle. The functions at this level will minimize human involvement and errors, enhance management efficiency, and facilitate the automated operation of a sustainable DC.

Through the roadmap, we aim to ultimately transform DC operations from the current reactive stage to an optimized and automated level. In what follows, we introduce relevant applications and enabling techniques.

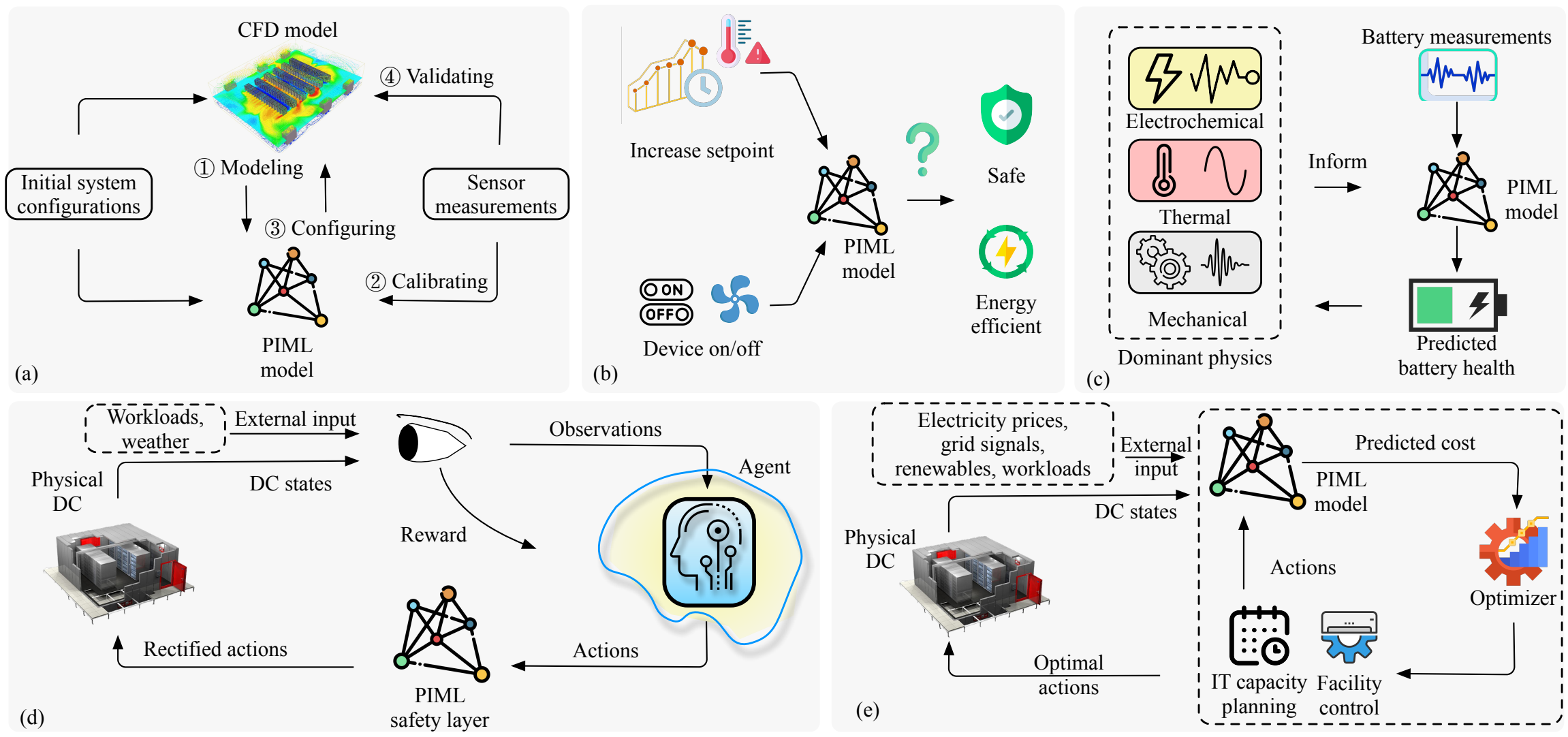

*Figure 5 MPIML-empowered applications across predictive and prescriptive levels. (a) CFD model auto-calibration, (b) cooling system what-if analysis, (c) UPS battery health prediction, (d) safety-aware intelligent cooling control, (e) joint IT-facility optimization with renewable and power grid integration.*

## 4. Applications & Enabling Technologies

This section provides examples of MPIML-empowered DC applications ranging from predictive to prescriptive levels.

### 4.1 Predictive Level Applications

**CFD Model Auto-Calibration**: CFD models have been widely used to prototype the data hall design and conduct what-if analysis during DC operations. However, manually calibrating the CFD models to identify proper parameter settings is labour-intensive. The PIML-based CFD surrogate models are advantageous as they can be continuously updated with online data and are differentiable concerning specific inputs, making them suitable for calibrating the original CFD model with unknown parameters. To illustrate, as shown in Figure 5 (a), the PIML-based model can serve as a surrogate of the original intensive and non-differentiable model, which can assist the parameter search via the gradient descent method. By integrating known physics and operational data, the physics-informed surrogate model strikes a balance between its accuracy and the overhead of producing training data from the original model. By ensuring the high fidelity of simulation models, this application provides the foundation needed for prescriptive optimizations to safely reduce energy consumption and the associated carbon footprint, moving beyond conservative, energy-intensive operational baselines.

**Cooling Failure Assessment**: Cooling system failures not only cause service breakdown but also trigger highly inefficient, energy-intensive emergency operations. Proactive assessment is therefore a prerequisite for running the facility at peak efficiency and minimizing energy waste as shown in Figure 5 (b). While various CFD software packages can be adopted, solving this model is computationally expensive and they do not take the associated plant failure into consideration. To address these issues, MPIML can be adopted to develop the transient data hall and chiller plant co-simulation model for cooling failure predictions. Specifically, the proper orthogonal decomposition (POD, which is a mathematical technique to simplify complex spatial data representation) with heat flux matching and physics-informed neural network (PINN, which is a neural network that embeds physics equations into its training process) have been studied to develop the air thermodynamics models. To couple the air thermodynamics with the chilled water loop, we need to model the heat exchangers, such as the cooling coils built within the CRAH. The heat transfer model calculates the water flow through a chilled water coil to meet the supply temperature and humidity ratio setpoints. It can be learned using implicit machine learning, i.e., finding the root of the water flow that satisfies the designed setpoints based on the energy balance equation. Additionally, the power consumption models of the facilities in the water loops, like pumps, chillers, and cooling towers, can be identified with the affinity laws and operational data. By incorporating prior knowledge, these models often require less data to train and exhibit better extrapolation capabilities compared with pure data-driven models.

**Battery System Health Forecasting**: The uninterrupted power supply (UPS) system is critical for reliability, but its batteries are also essential for integrating on-site renewable energy and participating in grid-level demand response. Forecasting battery health is crucial not only to prevent downtime but also to maximize the lifespan of battery assets for amortizing embodied carbon while ensuring storage is available when intermittent renewables are offline. To avoid risks and reduce shutdown costs, it is imperative to forecast the state of health (SoH) and remaning useful life (RUL) of the UPS batteries and take proper maintenance in advance. However, the UPS batteries often work in a floating state that is rarely discharged and charged, providing less dynamic data to develop health prediction models. To address this challenge, MPIML, which can holistically leverage the battery's internal electrochemical, mechanical, and thermal processes, holds great potential to tackle the data scarcity issue as shown in Figure 5 (c). The battery health is highly related to the impedance spectrum, which can be simulated via a circuit model that captures the battery's electrochemical properties. Additionally, the thermal and mechanical properties also serve as important anomaly indicators within batteries. For example, abnormal heat generation may indicate increased internal resistance or short circuits. The holistic applications of these physical processes will effectively supervise the health prediction model training and mitigate data demands.

## 4.2 Prescriptive Level Applications

**Safety-Aware Intelligent Cooling Control**: Cooling control optimization is important to reduce DC energy consumption and associated carbon footprint as cooling takes 35% electricity usage. Existing cooling controls often adopt the PID controllers to maintain the temperature and relative humidity at a target setpoint without considering energy usage. The prevalent DRL-based cooling optimization shows good performance in minimizing energy consumption and the associated carbon emissions. However, the DRL agents often learn from a trial-and-error process, which can pose potential risks. To mitigate these issues, it is important to supervise the agent's exploration with known physics. Specifically, as shown in Figure 5 (d), the thermodynamics models can be used as safety layers to guide the agent exploration by solving a quadratic programming. Under the physical constraints, the agent can learn an optimal cooling control policy with fewer safety violations. The explored data can also be used to identify parameters of specific facility energy models, such as CRAH fans, water pumps, and cooling towers. With the identified system models, we can adopt model-based DRL to obtain the energy-efficient cooling control policy.

**Renewable-Aware Joint IT-Facility Optimization**: With the growing DC electricity consumption, integrating renewable sources into the electricity supply is important to decarbonize a DC and achieve sustainable operations. However, renewable sources are often intermittent and unevenly distributed. To increase renewable utilization, one method is to jointly reschedule the IT workload and optimize cooling control. To jointly optimize load shifting and facility control, we need to develop models that can holistically assess the DC carbon footprint and forecast renewable generations. In this regard, MPIML is promising to achieve holistic DC carbon assessment by modeling the detailed facility energy consumption. With these models, as shown in Figure 5 (e), the optimization aims to reduce the overall DC carbon emissions, subject to SLA constraints and predicted demand, by adjusting cooling setpoints and provisioning IT resources. The optimization can be solved using MPC.

**Joint DC-Grid Optimization**: With the ever-growing capacity, DCs have become a major contributor to the power grid. Beyond on-site actions, DCs can act as large-scale flexible loads to support broader grid-level sustainability. By participating in demand response, a DC can reduce its consumption during peak times to reduce the grid's reliance on fossil-fuel plants and enable the grid to integrate more intermittent renewables without compromising stability. The DC can also regulate its energy consumption to participate in the demand response program, from which it can earn monetary revenues. Additionally, proper optimizations of the power system can improve the energy efficiency of the DC, e.g., by reducing the line loss in delivering electricity to data halls and the heat dissipation during the battery charging/discharging processes. To facilitate the optimization, our MPIML framework can model the holistic cost, including carbon emission cost, of DC and grid operations, allowing joint optimization to reduce both operational costs and the facility's grid-level carbon footprint as shown in Figure 5 (e). Joint optimization of the DC and power grid will reduce operational costs and boost overall sustainability.

## 4.3 Adaptive Level Applications

**Physics-Informed Lifelong Optimization**: The implementation of this stage relies on online learning and lifelong learning techniques. The physics-informed models are not static. They are designed for continuous recalibration. As new data from the physical DC becomes available, the framework automatically updates model parameters (e.g., thermal resistance, fan efficiency curves) to reflect system aging or hardware upgrades. This ensures that the optimization policies generated by DCBrain remain accurate and effective, minimizing human intervention and maintaining peak sustainability performance throughout the DC lifecycle.

## 5. Illustrative Example: MPIML-Empowered Cooling Control Optimization

This section presents an illustrative example on DC cooling control optimization to demonstrate the performance of the MPIML-empowered solution.

### 5.1 Testbed Introduction

To evaluate the performance of the MPIML-empowered solutions, we adopt the configurations and historical data of an industry-grade DC in Malaysia to build the simulation testbed. The DC consists of seven data halls, designed with an IT capacity of 18,350 kW. The chiller plant, as shown in Figure 6 (a), includes five chillers, each with a cooling capacity of 4,572 kW and a chilled water pump. Each condensed water loop includes two cooling towers operating in parallel, accompanied by a condensed water pump. In this case, we select the chiller plant and one associated data hall, as shown in Figure 6 (b), which consists of 22 CRAH units and hot aisle containment for evaluation. Specifically, the optimization aims to reduce the overall cooling system carbon emissions by adjusting facility setpoints, including the CRAH supply temperatures, the CRAH fan speed ratios, and the chilled water supply temperature. The SLA specifies that the IT equipment inlet temperature should be maintained within 27°C with relative humidity between 30% to 60%.

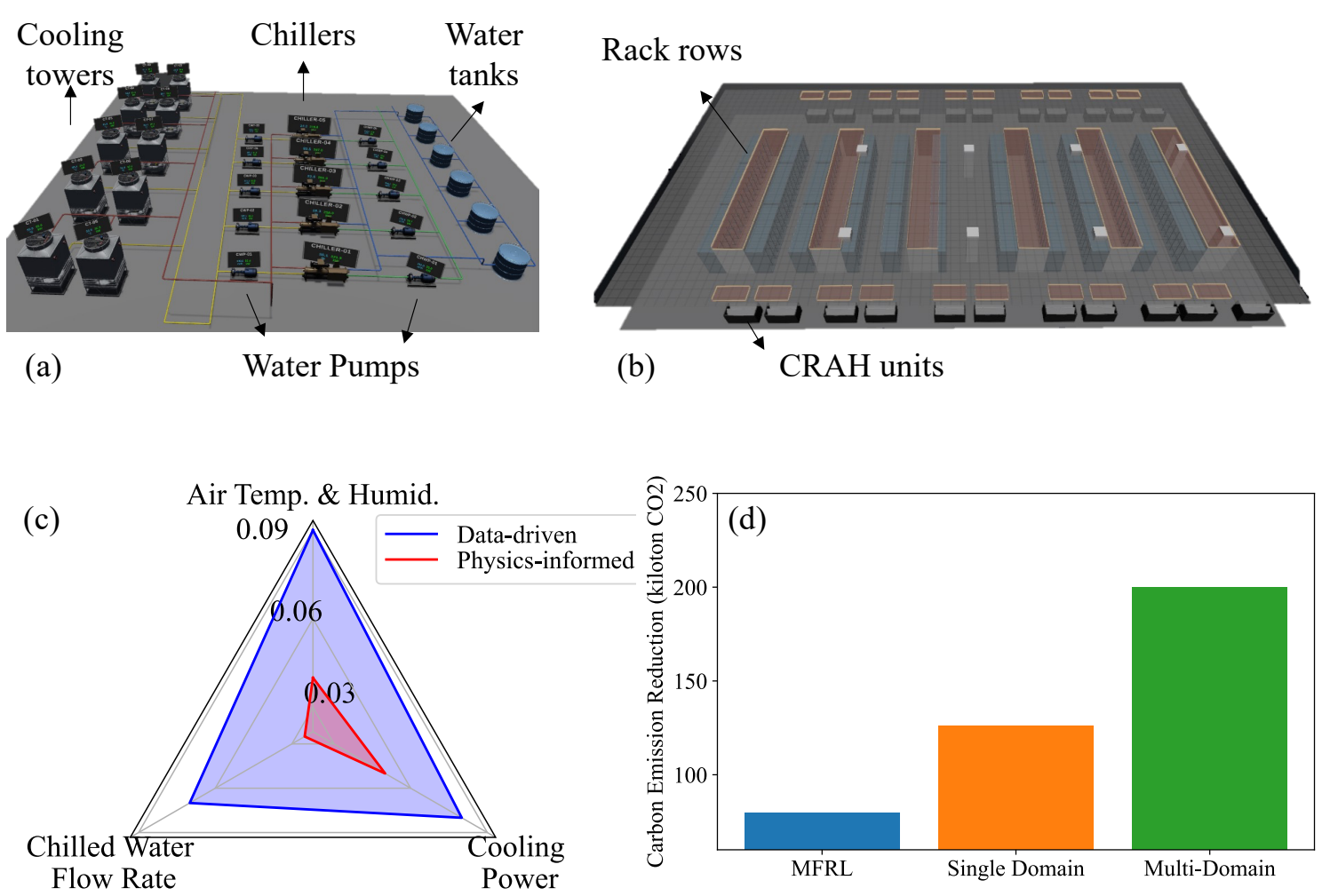

*Figure 6 Illustrative example. (a) Chiller plant testbed system, (b) one associated data hall for optimization, (c) mean relative model prediction errors that cover air, water, and electric power, and (d) carbon emission reductions of different methods compared with the existing policy over one year based on the grid emission factor of 0.758 kgCO2e/kwh in Malaysia.*

## 5.2 PIML Integration Mechanism

The PIML integration is realized in the coupled data hall and chiller plant system model. Specifically, we adopt the energy balance equation to model the data hall thermodynamics in a nodal form. The CRAH heat exchanger is modelled using the number of transfer units (NTU) method to calculate the heat transfer process between air and chilled water loop. For mechanical devices, the affinity law governs the power consumption like fan and pump. These equations are used as physical prior to develop the predictive model, whose unknown parameters (e.g., coil geometrical parameters, efficiency coefficients, performance curve coefficients,) were then learned from operational data. This physics-informed model provided the high-fidelity, differentiable environment used for the model-based optimization. The core principles of our MPIML framework are indeed applicable to other environments beyond tropical areas. For a colder region, the thermodynamic models would incorporate phenomena like free cooling. For emerging concepts like underwater data centers, the framework's models would focus on hydrodynamics and direct heat exchange with the surrounding water.

## 5.3 Comparison Methods

We evaluate the performance of model accuracy and carbon emission reductions for cooling system, respectively. For model accuracy, we compare the performance of pure data-driven and physics-informed models in predicting different physical variables, which include air temperature humidity, water flow rate, and power usage. For cooling power saving, we compare the performance of the following methods:

- **Fixed Policy**: This policy refers to the current DC operating by setting the supply temperatures, fan speed ratios, and chilled water to a fixed value. This is the current operating policy of the selected DC site. We will use it as the baseline for comparison of carbon emission reductions.
- **Model-Free RL**: This method adopts the soft actor-critic (SAC) as the single-agent optimization algorithm to optimize the control policy through extensive system interaction without building the dynamics models.
- **Single Domain Opt**: This method only considers the physics involved in the data hall, i.e., the thermodynamics and affinity laws, to develop data hall air-loop models for model predictive control without accounting for the chiller plant dynamics.
- **Multi-Domain Opt**: This method jointly considers the multi-physics involved in the data hall and chiller plant to develop holistic DC system models for optimization. Specifically, the air-water heat transfer, CRAH fan, chiller, and pump power usage are holistically considered.

For the model-based methods, i.e., single and multi-domain optimization, we used a predictive horizon of 1 hour with a control interval of 5 minutes. The optimization was solved using an open-source differentiable BFGS-SQP solver developed in PyGRANSO. For the model-free RL method, the SAC agent's policy and value function networks were composed of two hidden layers with 256 neurons each, using the ReLU activation function. The training batch size for the policy is set to 512 with an update frequency step of 288. The learning rate was set to 1e-3, with a replay buffer size of 1e6.

## 5.4 Performance Evaluation

First, we evaluate the model's prediction errors. Given the operational IT load and cooling control actions, the model aims to predict the return air temperature, relative humidity, chilled water flow rate required to meet the supply air setpoint, and the overall cooling system power usage, respectively. We select a week of DC operational data, using five days for training and the remaining two days for testing. Figure 6 (c) shows the radar chart of the test prediction errors, measured using the Mean Relative Error (MRE). This metric represents the absolute error divided by the ground truth of the observed values, making it a unitless percentage. For example, an MRE of 0.05 corresponds to a 5% average prediction error relative to the variable's ground truth. From the results, we observe that the physics-informed model, trained solely on operational data, generates lower MRE within 5%, compared with the data-driven model, which has higher MRE margins ranging from 7% to 9%.

Then, we adopt the trained model to optimize cooling control. Figure 6 (d) shows the annual carbon emission reductions with different optimization policies. Compared with the fixed baseline policy, we observe the MPIML-empowered policy achieves the highest carbon emission reductions of 200 kilotons per year, which is 2.5x and 1.6x of the model-free RL method and single domain optimization. This suggests the proposed approach can achieve holistic DC sustainability optimization and reach global optimal performance. Note that the performance of the model-free RL might be improved by introducing multi-agent control. However, without exploration supervision, the model-free RL often incurs severe SLA violations before the policy converges.

# 6. Challenges and Future Trends for Sustainable Operations

Our MPIML framework lays the groundwork for optimizing DC operations, but achieving the ultimate goal of fully autonomous, sustainable, and net-zero DCs requires addressing several key challenges. These challenges represent the next frontier in research and are critical for maximizing energy efficiency, enabling deep renewable integration, and ensuring safe, reliable operation as DCs evolve.

## 6.1 Model Adaptability and Scalability

A primary challenge is ensuring that PIML models remain accurate as DCs physically evolve. Current PIML models for thermodynamics are often geometry dependent, meaning they are trained for a specific physical layout. When new IT equipment is installed or air containment is modified, these models can become obsolete, leading to sub-optimal control and energy waste. Re-training them often requires extensive new CFD simulations, which is computationally expensive. This lack of adaptability is a direct barrier to long-term sustainability, as it prevents continuous optimization. Future work must focus on geometry-aware PIML, such as using Graph Neural Networks (GNNs), which can learn the underlying physics on unstructured meshes. Such models could adapt to layout changes on the fly, ensuring persistent energy efficiency and supporting the sustainable scalability needed to meet growing computing demands.

## 6.2 Heterogeneous and Multi-Scale Modeling

The rise of AI workloads is driving a shift toward heterogeneous hybrid cooling systems . These systems combine traditional macroscopic air cooling with microscopic, high-efficiency liquid cooling for powerful GPUs. From a sustainability perspective, this presents a critical multi-scale challenge, i.e., optimizing the chip-level liquid cooling must be done in synergy with the room-level air cooling. A failure to holistically model this system will negate the efficiency gains from liquid cooling, leading to significant energy waste and undermining decarbonization efforts. Future PIML frameworks must therefore be able to couple these different scales, for instance by integrating 1D fluid dynamic models for liquid loops with 3D models for airflow. This is essential for managing the high-power densities of AI-dedicated DCs sustainably.

## 6.3 Model Trustworthiness and Safety

Incorporating uncertainty quantification into the PIML framework promises significant advancements for mission-critical applications. Models are based on hypotheses and assumptions. This inherent hypothetical characteristic introduces a level of uncertainty into the models. As a result, this uncertainty also extends to the predictions they generate. Specifically, uncertainty may arise from various sources, including model assumptions, low-quality datasets, and the intrinsic variability of the system being modeled. These uncertainties can result in discrepancies between the model and its physical counterpart, potentially leading to less reliable predictions and suboptimal decision-making. Hence, this requires incorporating uncertainty quantification, which allows a model to not only make a prediction but also express its confidence in that prediction. By quantifying uncertainties arising from model assumptions or low-quality data, we can create safety-aware policies that optimize for energy efficiency while provably staying within safe operating limits. Building this trust is a non-negotiable step toward realizing the full energy-saving and carbon-reducing potential of intelligent DC operations.

# 7. Conclusion

The DC industry's escalating energy consumption presents a significant and growing challenge to global sustainability goals. To transition this critical digital infrastructure from a major energy consumer to a sustainable and flexible grid partner, we must move beyond simple efficiency and enable holistic sustainability optimization. This article proposes a MPIML-empowered framework designed to meet this challenge. By integrating fundamental physical laws with data-driven models, our approach provides the high-fidelity, reliable, and trustworthy digital twins necessary for mission-critical optimization. This method overcomes the safety and reliability concerns of traditional ML, creating a practical pathway for optimizing complex sustainability objectives, such as minimizing the carbon footprint and maximizing the integration of intermittent renewable energy sources. We presented an integrated system and an intelligence roadmap to advance DC operations from a reactive state to a fully prescriptive and adaptive one. Our illustrative example, based on an industry-grade facility, demonstrated tangible carbon emission reductions. This MPIML-based strategy is foundational to ensuring the digital backbone of our society can operate sustainably

## For Further Reading

Add up to six “For Further Reading” references at the end of each article. Self-citations by authors must be limited to no more than 2. Live links should be included whenever possible.

## Acknowledgements

This research is supported by Digitalland Innovation and Technology Pte. Ltd. under its Industrial Metaverse Platform for Data Center Management and Operations.

## Biographies

***Ruihang Wang, Qingang Zhang, Benjamin Jia and Yonggang Wen*** are with Nanyang Technological University, Singapore, 639798, SG

***Stuart Kennedy*** is with DayOne, Singapore, 038985, SG